\PassOptionsToPackage{prologue,dvipsnames}{xcolor}
\documentclass{article} 
\usepackage{iclr2025_conference,times}

\usepackage{amsmath,amsfonts,bm}

\def\eqref#1{equation~\ref{#1}}

\def\1{\bm{1}}

\DeclareMathAlphabet{\mathsfit}{\encodingdefault}{\sfdefault}{m}{sl}
\SetMathAlphabet{\mathsfit}{bold}{\encodingdefault}{\sfdefault}{bx}{n}

\usepackage{url}

\usepackage{morewrites}
\usepackage{bbm}
\usepackage{bm}
\usepackage{color}
\usepackage{multirow}
\usepackage{array}
\usepackage{booktabs}
\usepackage{diagbox}
\usepackage{bbding}
\usepackage{colortbl}
\usepackage{tabularx}
\usepackage{makecell}
\usepackage[dvipsnames]{xcolor}
\usepackage{hyperref}
\usepackage{subfigure}
\usepackage{amsmath, amssymb, overpic, textpos}
\usepackage{algorithmicx}
\usepackage{algpseudocode}
\usepackage{listings}
\usepackage{multicol}
\usepackage{xspace}
\usepackage{wrapfig}
\usepackage{graphicx}
\usepackage{graphbox}
\usepackage{arydshln}
\usepackage{algorithm}
\usepackage{adjustbox}
\usepackage{kotex}
\usepackage{caption}
\usepackage{setspace} 
\usepackage{epigraph}
\newcommand{\Ours}{{DeCo}\xspace}

\definecolor{uclablue}{rgb}{0.15, 0.45, 0.68}
\hypersetup{
    breaklinks,
    citecolor=uclablue,
    colorlinks=true,
}

\newlength\abovetabcapmargin
\newlength\belowtabcapmargin
\newlength\abovefigcapmargin
\newlength\belowfigcapmargin

\definecolor{SecondBest}{HTML}{FFFFA1}
\definecolor{Best}{HTML}{E8E800}

\newlength\savewidth

\newcommand{\upbad}[1]{\textcolor{Maroon}{\small \ $\uparrow${#1}}}
\newcommand{\up}[1]{\textcolor{PineGreen}{\small \ $\uparrow${#1}}}
\newcommand{\downbad}[1]{\textcolor{Maroon}{\small \ $\downarrow${#1}}}
\newcommand{\down}[1]{\textcolor{PineGreen}{\small \ $\downarrow${#1}}}

\newcolumntype{x}[1]{>{\centering\arraybackslash}p{#1pt}}
\newcolumntype{y}[1]{>{\raggedright\arraybackslash}p{#1pt}}
\newcolumntype{z}[1]{>{\raggedleft\arraybackslash}p{#1pt}}

\title{MLLM can see? Dynamic Correction Decoding for Hallucination Mitigation}

\author{
    Chenxi Wang$^{\spadesuit}$\footnotemark[1]~,
    Xiang Chen$^{\heartsuit}$\thanks{$\quad$ Equal Contribution.}~,
    Ningyu Zhang$^{\spadesuit}$\footnotemark[2]~,
    Bozhong Tian$^{\spadesuit}$,
    Haoming Xu$^{\spadesuit}$,\\
    \textbf{ Shumin Deng$^{\diamondsuit}$,
    Huajun Chen$^{\spadesuit\clubsuit}$\thanks{$\quad$ Corresponding Author.}}\\
    $^\spadesuit$Zhejiang University ~
     $^\heartsuit$Nanjing University of Aeronautics and Astronautics \\
    $^\diamondsuit$National University of Singapore, NUS-NCS Joint Lab, Singapore\\
    $^\clubsuit$Zhejiang Key Laboratory of Big Data Intelligent Computing\\
    \fontsize{10.2pt}{0.1\baselineskip}\selectfont \texttt{\{sunnywcx,zhangningyu\}@zju.edu.cn, {xiang\_chen}@nuaa.edu.cn}
}

\iclrfinalcopy
\begin{document}

\maketitle

\begin{abstract}
Multimodal Large Language Models (MLLMs) frequently exhibit hallucination phenomena, but the underlying reasons remain poorly understood. In this paper, we present an empirical analysis and find that, although MLLMs incorrectly generate the objects in the final output, they are actually able to recognize visual objects in the preceding layers. We speculate that this may be due to the strong knowledge priors of the language model suppressing the visual information, leading to hallucinations. Motivated by this, we propose a novel dynamic correction decoding method for MLLMs (\textbf{\Ours}), which adaptively selects the appropriate preceding layers and proportionally integrates knowledge into the final layer to adjust the output logits. Note that \Ours is model agnostic and can be seamlessly incorporated with various classic decoding strategies and applied to different MLLMs. We evaluate \Ours on widely-used benchmarks, demonstrating that it can reduce hallucination rates by a large margin compared to baselines, highlighting its potential to mitigate hallucinations\footnote{Code is available at \url{https://github.com/zjunlp/DeCo}.}.
\end{abstract}

\epigraph{\textit{``The first principle is that you must not fool yourself—and you are the easiest person to fool.''}}{--- Richard Feynman}

\section{Introduction}
\vspace{-5pt}

Recently, the rapid development of Multimodal Large Language Models (MLLMs) has demonstrated a potential pathway towards achieving Artificial General Intelligence (AGI) ~\citep{Qwen2-VL,minicpm,deepseek-vl,chameleon,gpt-4-tech,llava-1.5,DBLP:journals/corr/abs-2407-06135}.
However, in practice, the development of MLLMs is hindered by the phenomenon of hallucination, which typically results in the model generating statements about non-existent images while neglecting to mention certain visible objects, effectively causing it to fool itself~\citep{DBLP:journals/corr/abs-2404-18930,survey_lvlm,DBLP:conf/emnlp/LiDZWZW23,DBLP:journals/corr/abs-2306-14565,DBLP:journals/corr/abs-2309-05922}.
This issue poses significant risks in high-stakes fields such as medical imaging~\citep{huatuo-gptv,DBLP:journals/corr/abs-2304-04920,DBLP:journals/corr/abs-2302-07257}, autonomous driving~\citep{mllm_auto,drive_mllm}, and human-computer interaction systems~\citep{DBLP:journals/pacmhci/BrieBSV23}, where such errors could result in irreparable consequences.

The reasons behind hallucinations in MLLMs are complex. 
Unlike analyses focused on unimodal LLMs \citep{dola,DBLP:conf/icml/ChenXLWXGH24,orgad2024llmsknowshowintrinsic,chen2024llamaslayer8bshallow,lu2024insightsllmlongcontextfailures,surveyknow}, many current works assume that MLLM may indeed `see' visual information. 
However, due to factors such as excessive model depth \citep{DBLP:journals/corr/abs-2403-06764,Grad-CAM}, aggregation patterns \citep{opera}, or priors knowledge inherent in the MLLMs \citep{vcd,vdd}, these models ultimately still experience hallucinations (The details can be found in Appendix~\ref{sec:future}). 
Concretely, our understanding of the underlying mechanisms of hallucinations in MLLMs remains limited. 
It is still uncertain whether the visual information is never correctly recognized or if it is recognized but subsequently suppressed by later information streams.

\begin{figure*}[t!]
    \centering
    \subfigure[Object probing results.]{
        \includegraphics[width=0.6\textwidth]{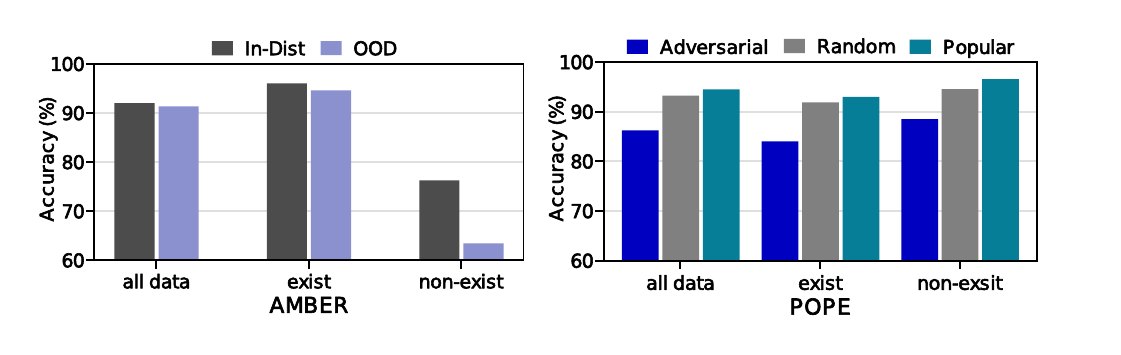}
        \label{fig:arc1}
    }
    \subfigure[Different resolution results.]{
        \includegraphics[width=0.35\textwidth]{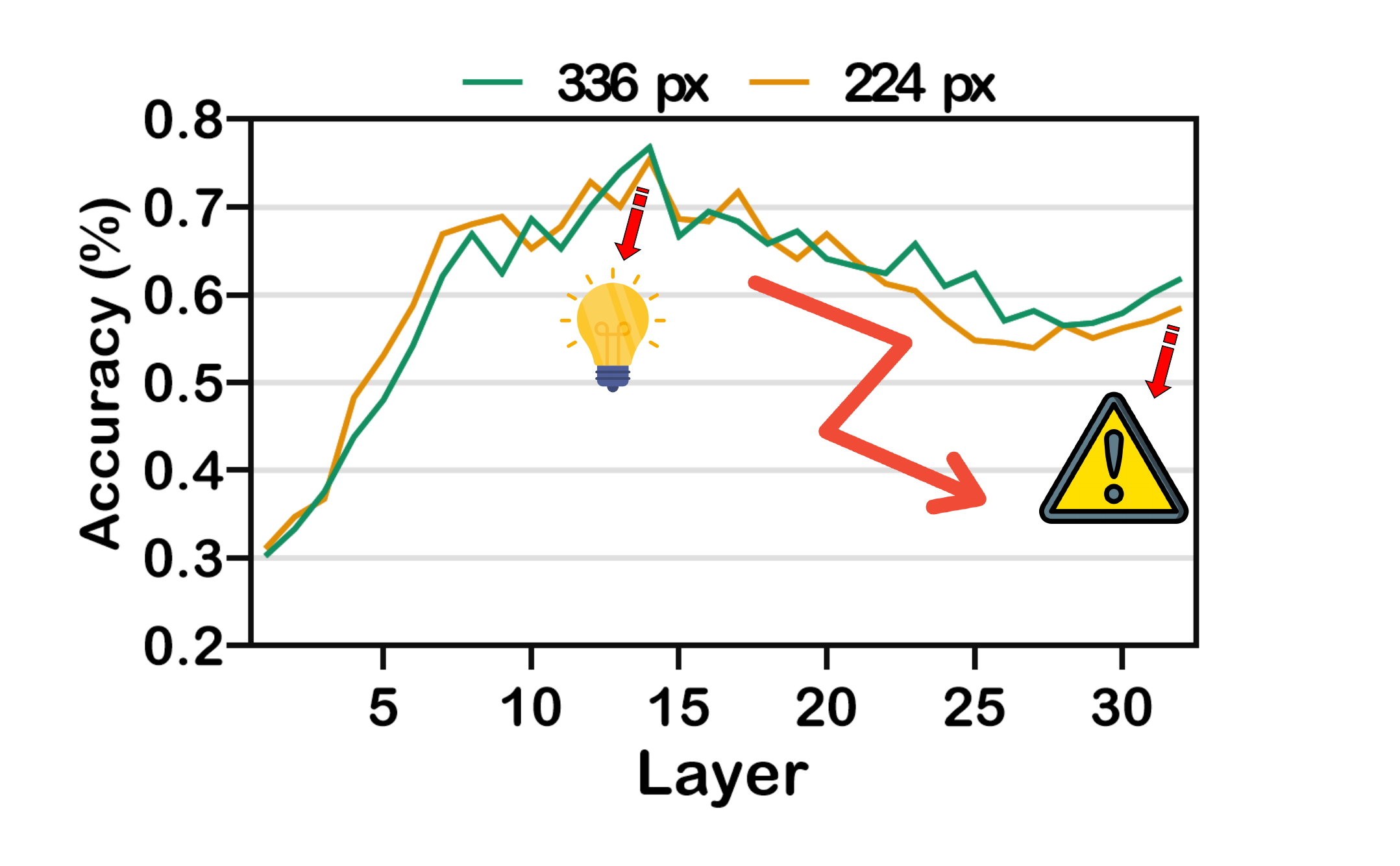}
        \label{fig:arc2}
    }
    \caption{Overall results of the probing experiment with MLLMs, indicating that they possess a certain level of awareness regarding the presence of visual objects (Figure~\ref{fig:arc1}), with prediction accuracy being higher in the preceding layers (Figure~\ref{fig:arc2}) but gradually \textbf{decline} afterward.}
    \label{fig:model}
\end{figure*}

\textbf{Hallucinated MLLM can see (to some extent).}  
Inspired by the aforementioned works, we conduct an empirical analysis and find that MLLMs are not blind; they can recognize objects in the preceding layers, but this recognition is suppressed in later layers, leading to hallucinations.
Specifically, we focus on object hallucinations\footnote{This approach is applicable to other types of hallucinations as well.} and conduct experiments with MLLMs, demonstrating that they know to some extent whether an object exists (as shown in Figure \ref{fig:model} and Section \ref{sec:finding1}).
We further observe that the confidence of generated tokens is influenced by the knowledge priors of MLLMs (Section \ref{sec:finding2}), leading to a reduction in the probability of ground truth tokens in the deeper layers.

\textbf{Dynamic correction decoding with preceding-layer knowledge.}
Based on those findings, we propose \textbf{D}ynamic Correction D\textbf{e}coding with pre\textbf{C}eding-Layer Kn\textbf{o}wledge (\textbf{\Ours}) to mitigate hallucinations for MLLMs.
Our core hypothesis is that preceding layers exhibit higher confidence for ground truth tokens, and the logits for these tokens should rank prominently at the last layer's outputs. 
To enhance the logits of ground truth tokens, \Ours dynamically selects preceding layer and utilizes its prior knowledge to correct the final output logits.
Additionally, we introduce a dynamic soft modulation to preserve the original style of the generated responses. 
\Ours is training-free and can be integrated with any popular decoding strategies, such as greedy search, nucleus sampling as well as beam search, and can seamlessly incorporate into any MLLMs for hallucination mitigation.

\textbf{Contributions.}
Our primary contribution lies in exploring the internal mechanisms of hallucinations in MLLMs. 
We find that the confidence of generated tokens is influenced by the knowledge priors of MLLMs, leading to a reduction in the probability of ground truth tokens in the deeper layers.
We further propose \Ours, a dynamic correction decoding method guided by preceding-layer knowledge. 
\Ours is integrated with InstructBLIP, MiniGPT-4, LLaVA, and Qwen-VL using three popular decoding strategies: greedy search, nucleus sampling, and beam search.
Experimental results show that \Ours achieves an average hallucination suppression rate of \textbf{10.8\%} in image captioning dataset, demonstrating superior suppression effectiveness. 
Additionally, \Ours outperforms baselines on visual question answering datasets including POPE, and MME.
Additionally, we analyze the latency and throughput, showing that \Ours introduces an approximate 1.2x increase in latency compared to the basic decoding process, much faster than previous baselines such as VCD and OPERA.


\section{Why do MLLMs generate non-exist objects?}
\label{sec:mech}

In this section, we conduct a series of  empirical analysis to investigate the internal mechanisms of MLLM and elucidate the underlying reasons for its generation of non-existent objects. To strike a balance between the realism and complexity of the experiments, we primarily focus on the generation of objects in image description scenarios (image caption tasks).
\paragraph{Preliminaries of MLLM generation.} 
MLLMs typically concatenate visual tokens, processed by the visual encoder and projection layer, with embedded textual tokens before feeding them into an autoregressive language model.
We denote the visual tokens as $\mathbf{X}^V = \{x_{v_1},x_{v_2},\ldots,x_{v_P}\}$ and textual tokens as $\mathbf{X}^C = \{x_{c_1},x_{c_2},\ldots,x_{c_Q}\}$. 
Here $P$ and $Q$ are the lengths of the visual tokens and textual tokens respectively.
Finally, the input is $\mathbf{X} = \text{concat}\{\mathbf{X}^V,\mathbf{X}^C\}$. 
Then $\mathbf{X}$ would be passed into MLLM with $\mathbf{N}$ stacked transformer layer. 
The intermediate variable generated by the i-th layer is called hidden states, denoted as $\mathbf{h}^i = \{h_0^i,h_1^i,\ldots,h_{T-1}^i\}$, where $T=P+Q$. 
During the generation phase, we use the hidden state at the last position in the final layer, which is mapped to the vocabulary dimension through an affine layer $\phi(\cdot)$, to predict the probability of the next token.
Formally, we have: 
\begin{equation}
\begin{aligned}
    p(x_T|x_{<T}) = \text{softmax}(\phi(h_{T-1}^N))_{x_T}, x_T\in \mathcal{V}
\end{aligned}
\end{equation}
where we use $x_{<T}$ to simplify the sequence $\{x_i\}_{i=0}^{T-1}$ and $\mathcal{V}$ refers to the whole vocabulary set.

\begin{figure}[t] 
\centering
\includegraphics[width=1\textwidth]{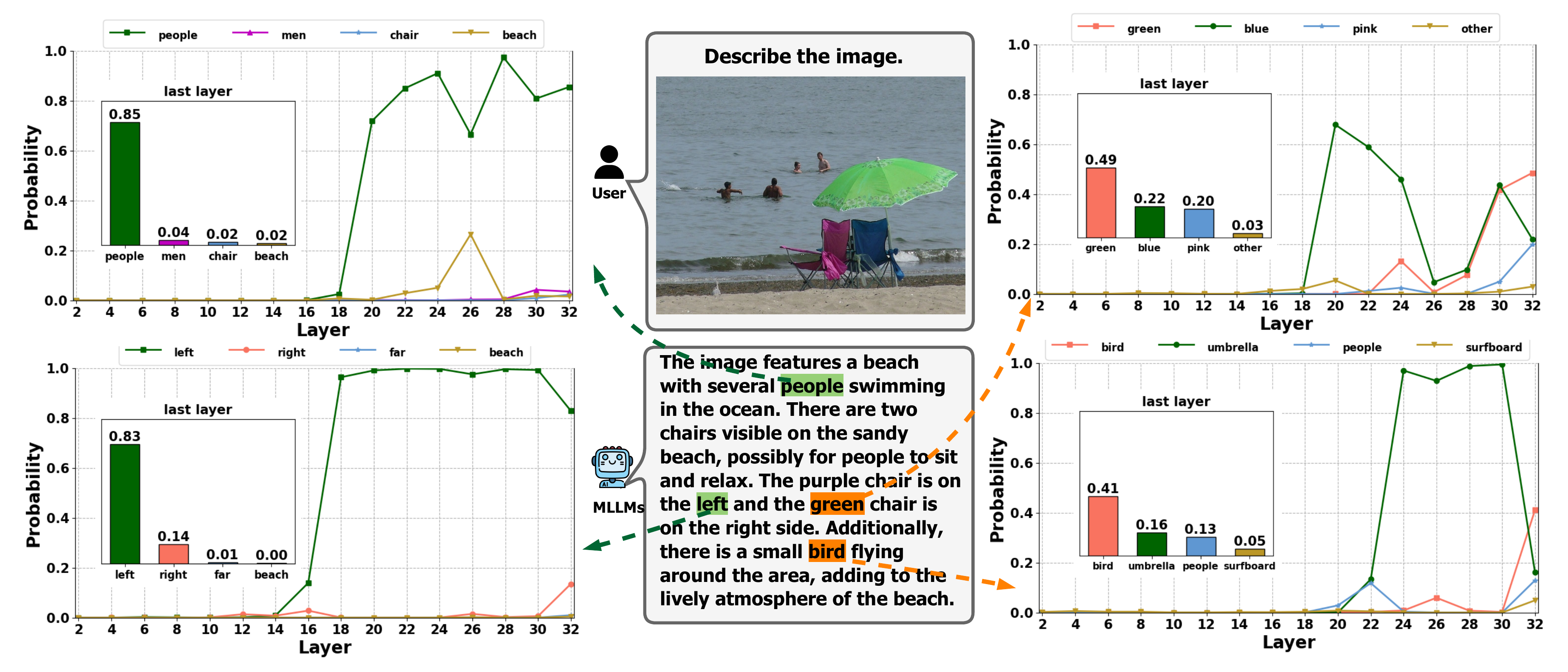}
\caption{
Illustration of token probabilities across transformer layers, which reveals distinct trends for target hallucinated (\textbf{\textcolor{BurntOrange}{orange}}) and non-hallucinated (\textbf{\textcolor{PineGreen}{green}}) tokens.
In the preceding layers, non-hallucinated tokens exhibit a higher probability.
In the final layers, hallucinated tokens demonstrate \textbf{increased probabilities}, while the probability of non-hallucinated tokens \textbf{drops sharply}.}
\label{fig:mec:case} 
\end{figure}

\subsection{Finding 1: MLLM knows to some extent whether an object exists}
\label{sec:finding1}

Inspired by \citep{DBLP:journals/corr/abs-2407-20311}, we explore how MLLMs comprehend objects in the image captioning task.
For simplicity, we abstract this process into a function called \textbf{isexist(obj)}, which determines whether an object is present in an image.
To examine the application of this function within the MLLM's image captioning workflow, we conduct probing experiments at the conclusion of object descriptions in each layer of the MLLM’s language model component, which consists of 32 transformer layers in a 7-billion-parameter model (Detailed setup in Appendix~\ref{appendix:exp1}).

We employ the prompt template, \textit{``USER: \textless image\textgreater Describe the image. 
ASSISTANT: The image contains {obj}.''}
Both the training and testing datasets are formatted accordingly before being input into MLLMs.
We train a probe classifier at the final position of the hidden state outputs for each transformer layer, resulting in a total of 32 classifiers. (For details on the subset division, OOD and in-distribution splits, and prompt templates, please refer to Appendix~\ref{appendix:exp1}.) The model is evaluated using the test set, as shown in Figure~\ref{fig:arc1} (left).
Further experiments are conducted on three splits of the evaluation dataset proposed by POPE, with results reported in Figure~\ref{fig:arc1} (right).
These evaluations provide a comprehensive understanding of the model's object recognition capabilities across diverse scenarios.

We select the best-performing probe classifier from the 32 classifiers to compare accuracy across all objects, existing objects, and non-existing objects.
Our results show that the MLLM achieves high accuracy for correctly generated objects in image captions.
Despite generating many non-existent objects, the MLLM still maintains around 80\% accuracy in our probing experiments.
This suggests that \textbf{MLLMs possess a certain level of understanding regarding object existence in images}.

Additionally, our probing experiments reveal higher accuracy in the preceding layers, as illustrated in Figure~\ref{fig:arc2}, which aligns with previous findings \citep{vdd,vcd}.
Furthermore, we show that increasing the resolution of the visual encoder (from 224px to 336px) enhances accuracy for non-existing objects, indicating that \textbf{token information at the last position in the preceding layers better represents visual information}. (For a detailed explanation of the different visual resolutions, please refer to Appendix~\ref{appendix:exp1}).
These findings suggests that the utilization of the preceding-layers in MLLMs enables the model to perform self-correction.

\subsection{Finding 2: Language Model Priors Suppress the Visual Information that MLLM Already See.}
\label{sec:finding2}

We hypothesize that the representations in the preceding layers effectively capture (to some extent) visual information. 
However, the prior knowledge embedded in the MLLM reduces the probabilities of ground truth tokens in deeper layers. Figure~\ref{fig:mec:case} illustrates this hypothesis with running examples. We analyze the Top-4 tokens ranked by probability in the final layer's output. Non-hallucinated tokens like \textit{``people''}, \textit{``left''}, \textit{``blue''}, and \textit{``umbrella''} exhibit high probabilities from the 18th layer. In contrast, hallucinated tokens like \textit{``bird''} and \textit{``green''} only show comparatively high probabilities around the 30-th layer. Interestingly, the probabilities of ground truth tokens \textit{``umbrella''} and \textit{``blue''} sharply decline from the 30-th layer onwards, eventually falling below the hallucinated tokens' probabilities in the final layer.


\begin{wrapfigure}{r}{0.35\textwidth}
\centering
\includegraphics[width=0.32\textwidth]{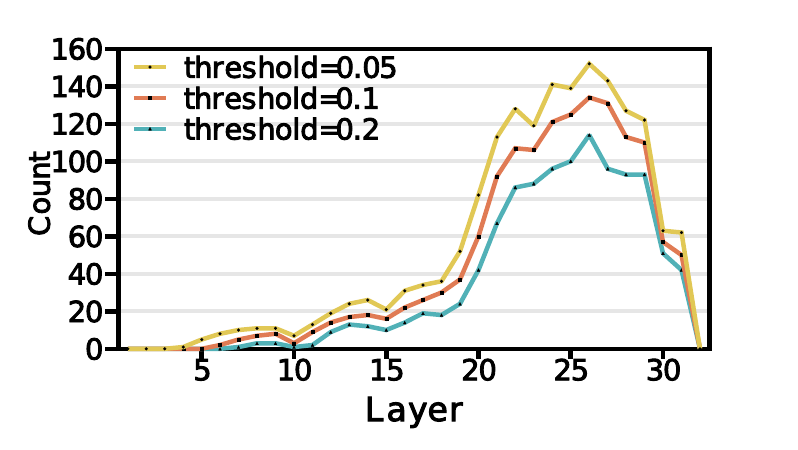}
\caption{Distribution of activated ground-truth tokens across layers.}
\label{fig:mech2}
\end{wrapfigure}

To further investigate this phenomenon, we conduct an early exit experiment ~\citep{DBLP:conf/icpr/Teerapittayanon16,DBLP:conf/iclr/ElbayadGGA20,DBLP:conf/nips/SchusterFG0B0TM22} to analyze the evolution of the MLLM's internal representations across transformer layers. We randomly select 500 images from the MSOCO dataset and use random prompts to elicit raw responses from LLaVA-1.5-7b. We then extract all non-existent objects along with their corresponding preceding text and input this data into the MLLM. We observe the probabilities of the next token across the transformer layers to gain insights into the model's behavior (see Appendix~\ref{appendix:exp2}  for detailed experimental setup).
The output of the $i$-th layer is denoted as $h^i$, and the probability distribution of the next token is represented as $p(\cdot|x_{<s})^i = \text{softmax}(\phi(h_{s-1}^i))$.
To reduce the observation tokens and simulate the real sampling process, we truncate the vocabulary,  similar to Top-$p$ sampling, and obtain the candidate tokens, denoted as $\mathcal{V}_{candidate}$ with a default threshold of 0.9. 
We then label the tokens in $\mathcal{V}_{candidate}$. Specifically, we filter out data where $\mathcal{V}_{candidate}$ contains at least one ground truth token and observe whether an \textbf{activated ground truth token} exists among the candidate tokens, formally expressed as:
\begin{equation}
\begin{aligned}
    \exists \ x_{a}\in \mathcal{V}_{candidate} \land i\in(0,N],\ p(x_{a}|x_{<s})^i - p(x_{h}|x_{<s})^i \geq \text{threshold},
\end{aligned}
\end{equation}
where $x_a$ is the activated ground truth token, $x_h$ is the token with the highest probability of being a hallucinated token in the probability distribution of the final layer and $\text{threshold}\in(0,1)$. Based on the experimental setup described above, we conducted the following investigation:

\paragraph{What suppresses the expression of visual facts?
 }
We analyze the occurrence of $x_a$ at each decoding layer, as shown in Figure~\ref{fig:mech2}. 
The results reveal that the activated ground truth tokens are primarily present between layers 20 and 28, indicating that MLLMs accurately recognize the image content in the latter layers. 
Notably, differences in experimental setups account for the variation in interval layers observed between Finding 1 and Finding 2.
However, the activated ground truth tokens are suppressed in the final output layer. This suppression may stem from the guidance of the input image or the inherent knowledge bias of the MLLM. To investigate this, we generate candidate tokens $\mathcal{V}_{candidate}^{'}$ in the absence of an input image, representing tokens based on the MLLM's inherent knowledge. We calculate that the overlap rate of $x_h$ existing in $\mathcal{V}_{candidate}^{'}$ reaches 91.05\%, suggesting that even without expressing image information, MLLMs still tend to generate the original hallucination tokens. This finding reveals that \textbf{the inherent knowledge in MLLMs may diminish the probability of the ground truth token in the deeper layers.}


\section{Proposed Approach: Dynamic Correction Decoding with Preceding-Layer Knowledge}

\begin{figure}[!ht] 
\centering
\includegraphics[width=1.0\textwidth]{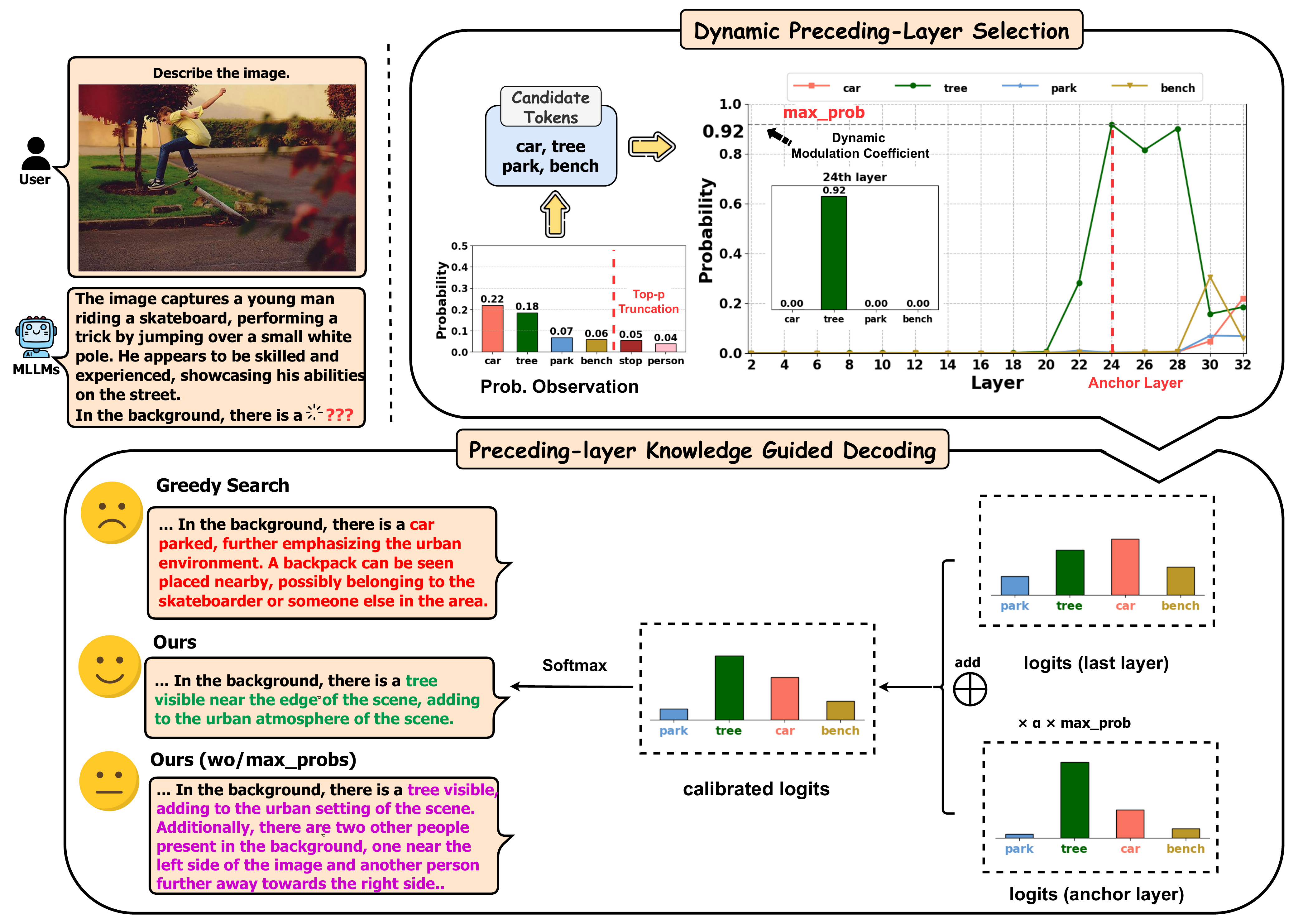}
\caption{Framework of {\Ours}. \Ours first dynamically selects an appropriate anchor layer from the preceding layers and then correct the knowledge in the final layer with dynamic coefficient.}
\label{framework} 
\end{figure}

After  investigating the reasons why MLLMs generate non-existent objects, inspired by \citep{dola}, we introduce \textbf{D}ynamic Correction D\textbf{e}coding with pre\textbf{C}eding-Layer Kn\textbf{o}wledge (\textbf{\Ours}), which can alleviate hallucinations during inference. 
The overall framework of Deco is illustrated in Figure~\ref{framework}, consisting of dynamic preceding layer selection (Section \ref{sec:m1}) and decoding correction with preceding-layer knowledge (Section \ref{sec:m2}).

\subsection{Dynamic Preceding-Layer Selection}
\label{sec:m1}
\paragraph{Candidate token acquisition.} 
Due to the vast vocabulary space, we track only the changes in the top-ranked tokens as candidate tokens across different layers for computational convenience.
This is based on the hypothesis that ground tokens usually appear in the top position of the MLLM's last layer output logits.
Inspired by \citep{DBLP:conf/acl/LiHFLEHZL23}, we use a truncation strategy to select the candidate tokens, with the default truncation strategy being top-$p$ truncation, formally:

\begin{equation}
\mathcal{V}_{\text {candidate}}\left(x_{T}|x_{<T}\right)=\left\{x_{T} \in \mathcal{V}: \sum_{v \in \mathcal{V}p} P_\tau(x_T = v | x_0, x_1, \dots, x_{T-1}) \leq p\right\}
\end{equation}

where $\mathcal{V}$ is the whole vocabulary, and $p$ refers to the parameter used in  top-$p$. 
The selected candidate tokens are theoretically ensured to be of high quality, thereby preventing the inclusion of low quality tokens (e.g.,semantically incorrect tokens) that exhibit high probabilities in preceding-layers but low probabilities in the final layer.

\begin{wraptable}{l}{0.35\textwidth}
\small
\centering 
\caption{Hit Rate of layers across different intervals.}
\begin{tabular}{lllc}
\toprule
\textbf{Layer Range} & \textbf{20-28} & \textbf{15-28} \\
\midrule
Hit Rate (\%) & \color{purple}{61.69}  & \color{purple}{71.14}  \\
\bottomrule
\end{tabular}
\label{tab:mech2}
\end{wraptable}

\paragraph{Preceding-layer selection.}
Our findings in Section~\ref{sec:mech} demonstrate that activated ground truth tokens typically exhibit higher probabilities in preceding layers compared to hallucinated tokens. Based on this observation, we hypothesize that selecting the token $x_{th}, \text{where}~x_{th}\in \mathcal{V}_{candidate}$, with the highest probability from the interval layers corresponds to the ground truth token. We compute the accuracy of $x_{th}$ as the ground truth token and denote this metric as the hit rate, as shown in Table~\ref{tab:mech2}. The results indicate that within a specific range of layers (e.g., 15-28), $x_{th}$ indeed has a high universal probability of representing the ground truth token. Intuitively, we track candidate tokens and dynamically choose the layer in which the token with the highest probability among the preceding layers resides to calibrate the final logit distribution of the MLLM. The selected preceding layer is referred to as the anchor layer, formally defined as:

\begin{equation}
    \mathcal{A}= \text{argmax}_i{\left\{x_{T} \in \mathcal{V}_\text{candidate}: \text{softmax}(\phi(h_{T-1}^i))_{x_T}, i\in \mathcal[a,b]\right\}},
\end{equation}
where $a\leq b, a,b \in [1,N]$, and $[a,b]$ represents the layer interval for MLLMs. Expanding the range of layers can improve the hit rate. To avoid increased search computation time, we assign default values of $a=20$ and $b=28$ for our subsequent experiments.

\subsection{Decoding Correction with Preceding-layer Knowledge}
\label{sec:m2}
\paragraph{Dynamic soft modulation.} We introduce a dynamic modulation coefficient, defaulting to the maximum probability. 
Formally, we have:
\begin{equation}
    \text{max\_prob} = \text{max}(\text{softmax}(\phi(h_{T-1}^\mathcal{A}))).
\end{equation}

This coefficient can help prevent hard changes in logits, particularly when the probability differences between candidate tokens in preceding layers are insignificant. 
From the example in Figure \ref{framework}, we can observe that the absence of the dynamic modulation coefficient may lead to semantic incoherence or even more severe hallucinations.

\paragraph{Preceding-layer knowledge guided decoding.} 
Given the selected preceding layers, we integrate information from these layers into the final layer to correct the logit distribution. 
We utilze a  hyperparameter, $\alpha$, to control the proportion of early-layer information incorporated. 
Additionally, dynamic soft modulation is employed to preserve the generative style of the original model. 
By utilizing the correction of preceding-layer representations, the probability of predicting the next token and the logits are updated as follows:
\begin{align}
& \hat{p}(x_{T} \mid x_{<T}) = \mathrm{softmax}\bigl(\text{logits}\bigr)_{x_T},\quad \\
& \text{logits} = \phi(h_{T-1}^N) + \alpha \times \text{max\_prob} \times \phi(h_{T-1}^\mathcal{A}),   
\end{align}
where $N$ is the last layer of MLLM and $\mathcal{A}$ is the selected preceding layer.






\section{Experiment}
\label{sec:exp}
\subsection{Setup}
\textbf{Baselines.} 
We integrate \Ours with various decoding methods, including greedy decoding, nucleus sampling, and beam search, and compare it against several baselines for mitigating hallucinations, as outlined below:
Dola~\citep{dola} is specifically designed for alleviating hallucinations in factual tasks for LLMs by reducing shallow semantic influences to improve the factuality of the final layer’s output. 
VCD~\citep{vcd} mitigates the influence of language model's priors in MLLMs by generating representations that enhance visual information through the subtraction of interfering knowledge prior during each sampling step.
OPERA~\citep{opera} dynamically penalizes overconfident tokens based on the emergence of aggregation patterns, while proposing a retrospective allocation strategy to avoid cases where hallucinations have already occurred. 
For all the baselines, we use the default hyperparameters from the source code for a fair comparsion.

\textbf{Model.} 
We select four of the most representative MLLM models for evaluation, including InstructBLIP~\citep{instructblip}, MiniGPT-4~\citep{minigpt4}, LLaVA-1.5 \citep{llava-1.5} and Qwen-VL ~\citep{Qwen-VL}. 
All the MLLMs we used have a language model size of 7 billion parameters (7B).

\textbf{Implementation Details.} 
To select the appropriate preceding layers for hallucination mitigation, we conduct ablation experiments, details of which can be found in the Section~\ref{sec:analysis}.
For a 7B-sized, 32-layer decoder-only architecture language model, we choose layers 20-28 as candidates for the preceding layers (according to the findings in Section \ref{sec:finding1}). For the image captioning and VQA tasks, $\alpha$ is set within the range of 0.1 to 0.6. In all experiments, we conduct inference on a single A800 GPU. The inference of 500 image-caption pairs take approximately 40 minutes.

\begin{table*}[t]            
\centering          
\caption{\textbf{CHAIR hallucination evaluation results}. 
Lower scores indicate fewer hallucinations. 
OPERA utilizes beam search, VCD applies nucleus sampling, and \Ours is the proposed method compatible with various decoding approaches. } 
\scalebox{0.68}{
\begin{tabular}{@{}llllllllll@{}}           
\toprule          
\multirow{2}{*}{\textbf{Decoding}} & \multirow{2}{*}{\textbf{Method}} &\multicolumn{2}{c}{\textbf{InstructBLIP}} &\multicolumn{2}{c}{\textbf{MiniGPT-4}} & \multicolumn{2}{c}{\textbf{LLaVA-1.5}} & \multicolumn{2}{c}{\textbf{Qwen-VL}}\\    
\cmidrule(lr){3-4} \cmidrule(lr){5-6} \cmidrule(lr){7-8} \cmidrule(lr){9-10}
  &  & $\mathrm{CHAIR_S}\downarrow$ & $\mathrm{CHAIR_I}\downarrow$ & $\mathrm{CHAIR_S}\downarrow$ & $\mathrm{CHAIR_I}\downarrow$ & $\mathrm{CHAIR_S}\downarrow$ & $\mathrm{CHAIR_I}\downarrow$ & $\mathrm{CHAIR_S}\downarrow$ & $\mathrm{CHAIR_I}\downarrow$ \\            
\midrule            
\multirow{2}{*}{\makecell[l]{Greedy}} & Vanilla & 58.8 & 23.7 & 31.8 & 9.9 & 45.0 & 14.7 & 46.0 & 12.5\\      
& DoLa & 48.4 & 15.9 & 32.2 & 10.0 & 47.8 & 13.8 & 46.8 & 12.9\\   
 & \textbf{\Ours (Ours)} & \textbf{41.2\down{17.6}} & \textbf{14.4\down{9.3}} & \textbf{27.0\down{4.8}} & \textbf{8.8\down{1.1}} & \textbf{37.8\down{7.2}} & \textbf{11.1\down{3.6}} & \textbf{42.2\down{3.8}} & \textbf{10.7\down{1.8}} \\        
\cmidrule(l){1-10}      
\multirow{3}{*}{\makecell[l]{Beam Search}} & Vanilla & 55.6 & 15.8 & 30.6 & 9.5 & 48.8 & 13.9 & 41.8 & 10.8\\            
 & OPERA & 46.4 & 14.2 & 26.2 & 9.5 & 44.6 & 12.8 & 34.6 & 9.5\\           
 & \textbf{\Ours (Ours)} & \textbf{43.8\down{11.8}} & \textbf{12.7\down{3.1}} & \textbf{24.8\down{5.8}} & \textbf{7.5\down{2.0}} & \textbf{33.0\down{15.8}} & \textbf{9.7\down{4.2}} & \textbf{32.0\down{9.8}} & \textbf{8.7\down{2.1}} \\          
\cmidrule(l){1-10}      
\multirow{3}{*}{\makecell[l]{Nucleus}} & Vanilla & 54.6 & 24.8 & 32.6 & 10.7 & 48.8 & 14.2 & 49.2 & 13.1\\            
 & VCD & 58.0 & 17.0 & 33.8 & 11.1 & 54.0 & 16.0 &  46.4 & 11.9\\           
 & \textbf{\Ours (Ours)} & \textbf{43.6\down{11.0}} & \textbf{12.9\down{11.9}} & \textbf{30.8\down{1.8}} & \textbf{9.5\down{1.2}} & \textbf{42.8\down{6.0}} & \textbf{13.2\down{1.0}} & \textbf{43.8\down{5.4}} & \textbf{11.8\down{1.3}}\\         
\bottomrule          
\end{tabular}}
\label{tab:chair}            
\end{table*}

\subsection{Benchmark and metrics}
\noindent\textbf{CHAIR.} Caption Hallucination Assessment with Image Relevance (CHAIR)~\citep{chair} metric, widely used in image captioning, identifies hallucinated objects by comparing the extracted objects with ground truth labels and evaluates both at the instance level ($\mathrm{CHAIR_I}$) and sentence level ($\mathrm{CHAIR_S}$), as shown in Eq.~\ref{eq:chair}. 
Following~\citep{opera}, we conduct experiments using the same settings, including the consistent 500 images from the MSCOCO 2014 validation dataset and the identical prompt, ``\texttt{Please help me describe the image in detail.}''.
\begin{align}  
\label{eq:chair}
    \mathrm{CHAIR_I} = \frac{|\{\text{hallucinated objects}\}|}{\text{all mentioned objects}}, 
    \mathrm{CHAIR_S} = \frac{|\{\text{captions with hallucinated objects}\}|}{\text{all captions}}.   
\end{align}

\begin{table*}[t]            
\centering          
\caption{\textbf{POPE hallucination evaluation results}. The best results are in bold.} 
\scalebox{0.72}{
\begin{tabular}{@{}llllll@{}}           
\toprule          
\multirow{2}{*}{\textbf{Decoding}} & \multirow{2}{*}{\textbf{Method}} &\multicolumn{1}{c}{\textbf{InstructBLIP}} &\multicolumn{1}{c}{\textbf{MiniGPT-4}} & \multicolumn{1}{c}{\textbf{LLaVA-1.5}} & \multicolumn{1}{c}{\textbf{Qwen-VL}}\\    
  &   & $\mathrm{F1}\uparrow$  & $\mathrm{F1}\uparrow$  & $\mathrm{F1}\uparrow$  & $\mathrm{F1}\uparrow$ \\            
\midrule            
\multirow{3}{*}{\makecell[l]{Greedy}} & Vanilla  & 80.0  & 58.5  & 82.2  & 85.2 \\      
& DoLa  & 83.4  & 72.8  & 83.2  & 85.8 \\   
 & \textbf{\Ours (Ours)}  & \textbf{84.9\up{4.9}}  & \textbf{77.4\up{18.9}}  & \textbf{86.7\up{4.5}} & \textbf{86.3\up{1.1}} \\        
\midrule  
\multirow{3}{*}{\makecell[l]{Beam Search}} & Vanilla  & 84.4  & 70.3  & 84.9  & 85.3 \\            
 & OPERA  & 84.8 & 73.3 & 85.4  & 86.1 \\      
 & \textbf{\Ours (Ours)}  & \textbf{84.9\up{0.5}}  & \textbf{77.9\up{7.6}} & \textbf{86.7\up{1.8}}  & \textbf{86.4\up{1.1}} \\          
\midrule  
\multirow{3}{*}{\makecell[l]{Nucleus}} & Vanilla  & 79.8 & 52.8 & 83.1  & 84.5\\          
 & VCD  & 79.9 & 56.0 & 83.1  & 84.7\\  
 & \textbf{\Ours (Ours)}  & \textbf{81.8\up{2.0}}  & \textbf{63.8\up{11.0}} & \textbf{85.4\up{2.3}} & \textbf{85.2\up{0.7}}\\         
\bottomrule          
\end{tabular}}
\label{tab:pope}            
\end{table*} 
\noindent\textbf{POPE.} The Polling-based Object Probing Evaluation (POPE)~\citep{DBLP:conf/emnlp/LiDZWZW23} is a VQA-based metric for assessing object hallucination in MLLMs. 
It evaluates hallucinations by asking questions such as ``\texttt{Is there a \textless object\textgreater\space in the image?}'' where \textless object\textgreater\space is derived from three types of splits: random (randomly selected objects), popular (frequently occurring objects), and adversarial (objects closely related to those in the image). 
The evaluation includes 500 MSCOCO images, with six questions per image for each split. 
We use F1 score for performance evaluation.

\noindent\textbf{MME.} The comprehensive MLLM Evaluation benchmark (MME) ~\citep{fu2023mme} assesses the perceptual and cognitive abilities of MLLMs across a total of 14 subtasks, including tasks such as OCR, visual knowledge, attribute relationships, and object recognition. 

\noindent\textbf{GPT-4o assisted evaluation.}
To further assess the model's performance in image captioning, we extend beyond the CHAIR metric, which targets object hallucination. 
Following prior studies~\citep{opera,vcd}, an open evaluation is conducted using GPT-4o on 100 randomly sampled COCO images.
GPT-4o assesses two assistants' descriptions in terms of Accuracy (A) (e.g., truthfulness), Detailedness (D) (e.g., richness) and Coherence (C).
We introduce the prompt used in the experiments in Table~\ref{tab:4o-prompt} and Table~\ref{tab:coh}.

\subsection{Experimental Results}
\noindent\textbf{Results of hallucination in image captioning.}  
Note that we use the baseline's original decoding settings for a fair comparison and run \Ours under the same settings.
From Table~\ref{tab:chair}, we notice that \Ours consistently outperforms other approaches in mitigating hallucinations across four MLLMs—InstructBLIP, MiniGPT-4, LLaVA-1.5, and Qwen-VL—using three decoding strategies: greedy search, beam search, and nucleus sampling.
We find that \Ours slightly outperforms OPERA, while our method demonstrates higher efficiency and simplicity in inference (see Section \ref{sec:analysis}).
Additionally, VCD does not perform as well, likely due to producing an increased number of hallucinated descriptions during the generation process. 
In conclusion, the proposed approach \Ours effectively reduces hallucinations in visual description tasks solely through dynamic decoding correction, achieving an average suppression rate of approximately \textbf{10.8\%} on image captioning datasets.
Additionally, we further evaluate the performance of \Ours on the AMBER image caption dataset, as detailed in Table~\ref{tab:amber} of the Appendix.

\begin{wrapfigure}[11]{l}{0.36\textwidth}
\centering
\includegraphics[width=0.36\textwidth]{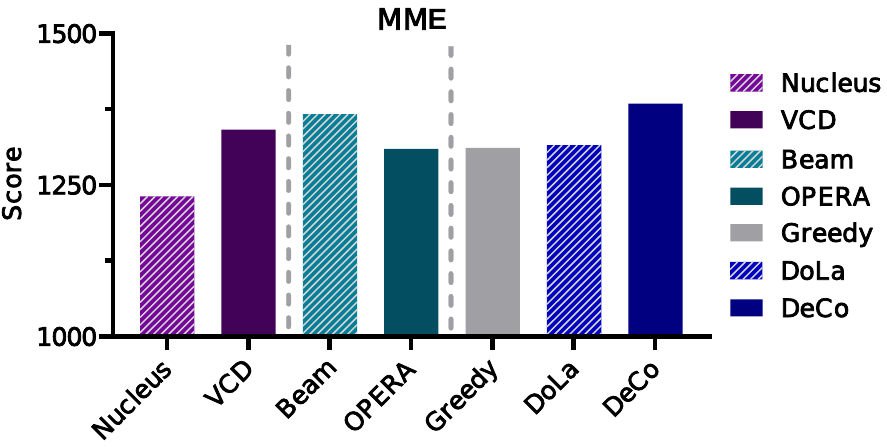}
\caption{\Ours generally improves the MLLM's performance.}
\label{fig:mme}
\end{wrapfigure}

\noindent\textbf{Results of hallucination in VQA.}
In contrast to image captioning, POPE employs a simple polling approach to assess hallucination levels in MLLMs with respect to object recognition.
As shown in Table~\ref{tab:pope}, \Ours demonstrates superior performance across all settings, further validating the effectiveness of the proposed approach.
Additionally, Figure~\ref{fig:mme} reveals that \Ours also achieves better results on MME, which evaluates the multifaceted VQA capabilities of LLaVA-1.5.
These findings suggest that the underlying mechanism we identified not only applies to object recognition but also extends to attribute-related tasks and more complex reasoning tasks.



\noindent\textbf{Results of GPT-4o's assistance.} 
Following~\citep{opera, vcd}, we further use GPT-4o to evaluate our method against greedy decoding across four distinct models.
From Table \ref{tab:gpt_4v}, we notice that our approach consistently outperform greedy decoding in terms of accuracy, demonstrating its efficacy in hallucination suppression. 
The impact of decoding intervention is evident in the level of detail produced: for some models, our method yield only marginally higher or, in certain cases, slightly lower levels of detail compared to greedy decoding.
\Ours also exhibits a coherence level comparable to that of the baseline.
Nonetheless, our method exhibit a clear advantage in mitigating hallucinations across all evaluated models.
\begin{table}[h]
\small
\centering 
\caption{GPT-4o assisted hallucination evaluation results on MSCOCO. Three aspects are verified, accuracy ($A$), detailedness ($D$) and coherence ($C$).}
    \footnotesize
    \centering
    \setlength{\tabcolsep}{1mm}{
    \begin{tabular}{lp{6mm}<{\centering}p{6mm}<{\centering}p{6mm}<{\centering}p{6mm}<{\centering}p{6mm}<{\centering}p{6mm}<{\centering}p{6mm}<{\centering}p{6mm}<{\centering}p{6mm}<{\centering}p{6mm}<{\centering}p{6mm}<{\centering}p{6mm}<{\centering}}
    
        \toprule
        \multicolumn{1}{c}{\multirow{2}{*}{Method}}
        & \multicolumn{3}{c}{InstructBLIP} 
        & \multicolumn{3}{c}{MiniGPT-4} 
        & \multicolumn{3}{c}{LLaVA-1.5} 
        & \multicolumn{3}{c}{Qwen-VL} 
        \\
        \cmidrule{2-4}
        \cmidrule{5-7}
        \cmidrule{8-10}
        \cmidrule{11-13}
        & $A$ & $D$ & $C$ & $A$ & $D$ & $C$ & $A$ & $D$ & $C$ & $A$ & $D$ & $C$ 
        \\
        \midrule
        Greedy Search & 4.92 & 5.65 & 6.89 & 5.71 & \textbf{6.20} &\textbf{7.67} & 5.21 & \textbf{6.31} &\textbf{8.18} & 5.56 & 6.62 & \textbf{8.20}
        \\
        \textbf{\Ours (Ours)} & \textbf{6.25} & \textbf{5.77} & \textbf{7.14} & \textbf{6.33} & 6.08 & 7.54 & \textbf{7.42} & 6.25 & 7.96 &\textbf{7.81} & \textbf{6.70} & 8.15
        \\
        \bottomrule
    \end{tabular}
    }
    \label{tab:gpt_4v}
\end{table}

\subsection{Analysis}
\label{sec:analysis}

\begin{wrapfigure}{r}{0.4\textwidth}
\vspace{-1.2cm}
\centering
\includegraphics[width=0.4\textwidth]{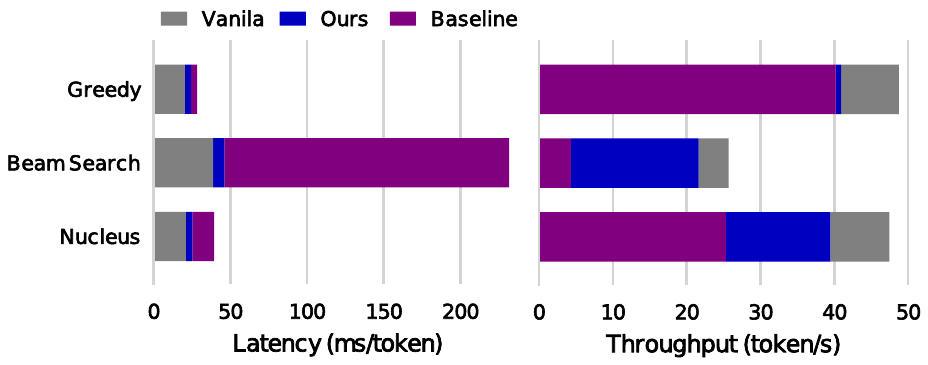}
\caption{Comparison of latency and throughput across different baselines.}
\label{fig:latency}
\vspace{-0.5cm}
\end{wrapfigure}

\paragraph{Latency and throughput analysis.}
To evaluate the efficiency of \Ours, we compare its latency and throughput with several baselines, \textbf{including DoLa, OPERA, and VCD based on Greedy, Beam Search, and Nucleus Sampling, respectively}. Figure~\ref{fig:latency} illustrates the results of this comparison. 
The findings indicate that {\Ours} operates within an acceptable efficiency cost, striking a balance between effectiveness and computational overhead. 
Compared to the basic decoding process, the latency increase introduced by our method is approximately 1.2 times.
In contrast, the latency increases for VCD and OPERA are 1.8 and 5.1 times, respectively.
While both VCD and OPERA demonstrate comparable efficacy in mitigating hallucinations, their computational overheads remain relatively high. 
This highlights the practical value of  {\Ours}, as it can be integrated into real-world applications without significantly compromising efficiency.

\paragraph{Perturbation in the selected preceding-layer.} 
To evaluate the effectiveness of the dynamic layer selection method, we introduce a random perturbation strategy. 
Specifically, for the predetermined preceding layers, we add random values ranging from -5 to 5 to modify the selection of layers. 
We randomly select 200 images from the MSCOCO dataset and prompt MLLMs to generate descriptions. 
The results after incorporating the perturbations are presented in Table~\ref{tab:pertubation}. 
Notably, the perturbed results demonstrate a significant degradation in performance, further validating the effectiveness of our proposed method.

\begin{table}[!ht]
\caption{Comparison of results between \Ours and perturbed \Ours in image captioning tasks}
    \footnotesize
    \centering
    \scalebox{0.8}{
    \begin{tabular}{lllllllll}
    
        \toprule
        \multicolumn{1}{c}{\multirow{2}{*}{Method}}
        & \multicolumn{2}{c}{InstructBLIP} 
        & \multicolumn{2}{c}{MiniGPT-4} 
        & \multicolumn{2}{c}{LLaVA-1.5} 
        & \multicolumn{2}{c}{Qwen-VL} 
        \\
        \cmidrule{2-3}
        \cmidrule{4-5}
        \cmidrule{6-7}
        \cmidrule{8-9}
        &  $\mathrm{CHAIR_S}\downarrow$ & $\mathrm{CHAIR_I}\downarrow$ & $\mathrm{CHAIR_S}\downarrow$ & $\mathrm{CHAIR_I}\downarrow$ & $\mathrm{CHAIR_S}\downarrow$ & $\mathrm{CHAIR_I}\downarrow$ & $\mathrm{CHAIR_S}\downarrow$ & $\mathrm{CHAIR_I}\downarrow$
        \\
        \midrule
        \Ours & 39.3 & 12.6 & 32.4 & 9.6 & 38.8 & 11.1 & 44.5 & 11.1
        \\
        \Ours + $\epsilon$ & 45.6\textbf{\upbad{6.3}} & 14.3\textbf{\upbad{1.7}} & 33.3\textbf{\upbad{0.9}} & 10.1\textbf{\upbad{0.5}} & 42.2\textbf{\upbad{2.4}} & 11.3\textbf{\upbad{0.2}} & 47.0\textbf{\upbad{2.5}} & 12.8\textbf{\upbad{1.7}}
        \\
        \bottomrule
    \end{tabular}
    }
    \label{tab:pertubation}
\end{table}

\begin{figure*}[h]
    \centering
    \subfigure[Ablation study of $\alpha$.]{
        \includegraphics[width=0.44\textwidth]{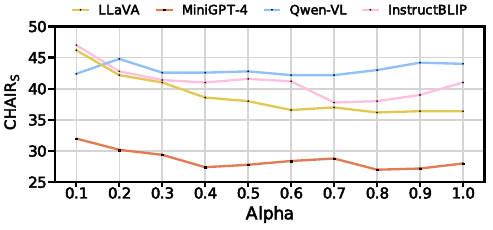}
        \label{fig:alpha}
    }
    \subfigure[Ablation study of interval layers.]{
        \includegraphics[width=0.44\textwidth]{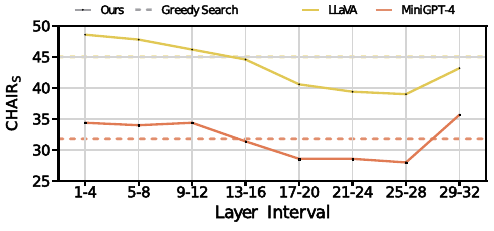}
        \label{fig:layer_range}
    }
    \caption{Ablation experiment results for hyperparameter $\alpha$ and different interval layers.}
    \label{fig:ablation}
\end{figure*}

\paragraph{Hyperparameter analysis.} 
Our method incorporates two primary hyperparameters: $\alpha$ and the selection of interval layers. 
In the experiments, we employ \Ours based on greedy decoding.
On the one hand, the hyperparameter $\alpha$ regulates the intensity of early information enhancement.
Figure~\ref{fig:alpha} illustrates the performance across various $\alpha$ values.
We observe that hallucination suppression is most effective when $\alpha$ approximates 0.6.
As $\alpha$ increases, the efficacy of \Ours in mitigating hallucinations improves.
However, it is crucial to note that excessively high $\alpha$ values may lead to the generation of atypical image descriptions, characterized by repetitive word usage. 
Notably, we can adjust the value of alpha appropriately to balance the truthfulness and semantic coherence of the responses (e.g., by using lower alpha). Additionally, our approach and the hyperparameter for repetition penalty are orthogonal, which implies that we can introduce the repetition penalty term to mitigate repetition.
On the other hand, the layer interval hyperparameter $\mathcal[a,b]$ determines the candidate layers for inclusion in the enhancement process. 
We conduct experiments using intervals of four layers, with results presented in Figure~\ref{fig:layer_range}. 
Our analysis reveals that hallucination suppression for MLLM is negligible in layers 1-16, while layers 20-28 demonstrate substantial mitigation of hallucinations.
Notably, layers 29-32 exhibit minimal hallucination suppression, aligning with our findings discussed in Section~\ref{sec:finding2}.
For other families of MLLMs and larger scale MLLMs, the selection of interval layer should be appropriately adjusted based on empirical experimentation.

\paragraph{Mitigating snowballing hallucinations.} Snowballing hallucinations are a prevalent issue in the responses generated by MLLMs.
This phenomenon occurs when an initial hallucination triggers a sequence of subsequent errors, leading to a compounding effect that significantly degrades the quality and coherence of the generated text. Figure~\ref{snowballing} illustrates a typical example of snowballing hallucinations, where an initial misinterpretation of the visual input propagates through the decoding process, resulting in a highly inconsistent and erroneous output.
Our approach can reduce the accumulation of errors and improves the overall consistency and accuracy of the generated responses. 
The effectiveness of \Ours is further demonstrated through additional cases based on diverse MLLMs, which can be found in Figures~\ref{fig:blip_cases}, \ref{fig:llava_cases}, \ref{fig:qwen_cases}, and \ref{fig:minigpt_cases} in Appendix~\ref{appendix:case}.

\section{Related Work}

\subsection{MLLM Hallucination Mechanism}
Hallucination in MLLMs, characterized by contradictions between image input and textual output, has been a prevalent issue \citep{survey_lvlm,unihd}. 
Current research on the mechanism of hallucination in MLLMs focuses on two key aspects: the interaction between images and text at different layers, and the prior bias of the LLM during decoding.
Several studies have investigated the role of image-text interaction at different layers in MLLMs. 
Grad-CAM \citep{Grad-CAM} visualizations reveal that image-text interaction exists in the preceding layers (1-11) but not in the deep layers.
OPERA \citep{opera} further proposes that the ``Aggregation Pattern'' leads to hallucination, where visual information from preceding layers is gradually aggregated to anchor tokens, and focusing solely on these tokens during prediction while ignoring visual information leads to a high probability of hallucination in the generated sequence.
However, other studies have revealed that MLLMs exhibit biases towards LLM priors, even in the presence of noisy or absent visual information.
VCD \citep{vcd} discovers that MLLMs generate high-confidence answers even when the image is noisy or absent, indicating a bias towards LLM priors.  
Similarly, PAI \citep{liu2024payingattention} describes this phenomenon as ``Text Inertia'' and posits that it stems from existing paradigms that map visual representations onto the text representations as tokens. 
This leads to an inference process that fails to adequately account for image tokens, resulting in hallucinations.

\subsection{Hallucination Mitigation for MLLMs}
One straightforward approach to mitigate hallucination is to reduce the knowledge gaps and data bias between vision and language during model training. 
Finetuning-based methods have been explored, focusing on crafting specific datasets \citep{ferret,DBLP:conf/aaai/GunjalYB24,ijcai2024p0687} and alignment training \citep{LLaVA-RLHF,RLHF-V,archit1,archit2} to achieve better knowledge alignment between images and text. While these methods have shown promising results, they often require expensive annotated paired data and substantial computational resources.

Hallucination can also be mitigated by post-processing methods, which usually involve using additional tools or self-reflection strategies to revise the response.
For instance, LURE \citep{lure} detects hallucinations using manually-crafted features and revises the generated text accordingly. 
Woodpecker \citep{woodpecker} combines MLLM outputs with an expert VQA model to post-edit hallucinations. 
However, these methods incur additional inference costs and delays, and require task-specific procedures and prompts to be designed \citep{inevitable}.
Training-free decoding methods have been explored to mitigate hallucination. 
OPERA \citep{opera} identifies an abnormal attention pattern that often accompanies hallucinated descriptions and proposes the mitigation method based on this pattern. VCD \citep{vcd} introduces the notion that visual uncertainty increases hallucination and proposes a contrast decoding method to alleviate the issue. 
VDD \citep{vdd} proposes a ``Post-Hoc debias'' approach that ensures uniform scores for each answer in the absence of an image to mitigate the influence of LLM priors. 

\section{Conclusion}

In this paper, we demonstrate that MLLMs exhibit an awareness of hallucinated objects, with earlier layers showing higher confidence, while tokens shaped by prior knowledge diminish the likelihood of true tokens in the final layers.
Based on this insight, we introduce \Ours, dynamic correction decoding with preceding-layer knowledge to mitigate hallucinations. 
Extensive experiments demonstrate the efficacy of our approach, which also shows  advantages in latency and throughput.

\section*{Acknowledgments}
This work was supported by the National Natural Science Foundation of China (No. 62206246, No. NSFCU23B2055, No. NSFCU19B2027), the Fundamental Research Funds for the Central Universities (226-2023-00138), Yongjiang Talent Introduction Programme (2021A-156-G), CIPSC-SMP-Zhipu Large Model Cross-Disciplinary Fund, Ningbo Science and Technology Special Projects under Grant No. 2023Z212, Information Technology Center and State Key Lab of CAD\&CG, Zhejiang University, NUS-NCS Joint Laboratory (A-0008542-00-00), and the Ministry of Education, Singapore, under the Academic Research Fund Tier 1 (FY2023) (Grant A-8001996-00-00).
We gratefully acknowledge the support of Zhejiang University Education Foundation Qizhen Scholar Foundation.

\section*{Reproducibility statement} 
We have submitted the relevant code in the supplementary materials. 
The names of the experimental benchmarks, the prompt templates used, and the model's hyperparameter settings can all be found in Section~\ref{sec:exp}. 
The Appendix~\ref{appendix:exp1} and \ref{appendix:exp2} provides a detailed description of the experimental setup for the mechanism experiments.



\bibliography{iclr2025_conference}
\bibliographystyle{iclr2025_conference}

\appendix
\appendix
\section{Limitations}
\paragraph{Lack of generalized research.}
Due to the GPU cost consideration, we conduct experiments solely on limited MLLMs, without exploring additional MLLMs or those with larger parameter sizes.
\paragraph{No free lunch.}
The results shown in Table~\ref{tab:gpt_4v} indicate that our method has a little negative impact on the level of detailedness metric.
In future work, we aim to integrate \Ours with other strategies and explore approaches that can effectively balance truthfulness and diversity.

\section{Comparative Analysis, Summary, and Future Directions}
\label{sec:future}

Here, we compare our work with previous works, summarize and speculate on the underlying causes of hallucinations in LLMs and MLLMs, and provide insights into future directions.
\paragraph{Comparison of previous mechanism findings.}
Existing studies suggest that MLLMs may focus more on visual tokens in the early layers while paying greater attention to textual tokens in the later layers \citep{Grad-CAM,DBLP:journals/corr/abs-2403-06764}. 
The aggregation pattern is typically positively correlated with hallucinations and tends to emerge at deeper layers \citep{opera}. 
These conclusions align with our findings, suggesting that MLLMs exhibit a better ability to perceive visual information in the preceding-layers compared to the final layers. 
In the detection of hallucinations in LLMs, some studies employing probing techniques have found that the intermediate layers exhibit the best detection performance \citep{inside,orgad2024llmsknowshowintrinsic,chen2024llamaslayer8bshallow,lu2024insightsllmlongcontextfailures}, a finding similar to our Finding 1. This suggests that the hallucination mechanisms in LLMs and MLLMs may share underlying similarities.
\paragraph{Comparison of previous hallucination mitigation methods.}
Our work shares a similar assumption with OPERA \citep{opera} and VCD~\citep{vcd}, positing that the knowledge priors inherent in MLLMs may suppress the model's ability to comprehend visual information. However, our approach is comparatively simpler than that of OPERA \citep{opera} and VCD~\citep{vcd}.
Additionally, our work differs from the assumption in unimodal LLMs, where the semantic information present in the shallow layers interferes with factual recall in the final layer \citep{dola,DBLP:conf/icml/ChenXLWXGH24}.
However, our method is actually parallel to previous approaches and can be combined to achieve better results.
\paragraph{Summary.}
Combining current research, we speculate that this phenomenon observed in both LLMs and MLLMs may be due to characteristics of the Transformer architecture, specifically the anchor token effect in the attention mechanism \citep{opera,anchor}, which leads to information loss when processing long sequences. For instance, in MLLMs, a single token may be insufficient to summarize information from extended sequences of visual tokens.
Another work suggest that the knowledge overshadowing of multiple conditions within the query leads to hallucinations in LLM \citep{overshadow}. In multimodal settings, image information represents a distinct condition. When the textual modality overshadows the condition related to the image, it can result in hallucinations in visual perception. Essentially, this reflects a loss of information flow within the attention mechanism. 
Overall, from an architectural perspective, hallucinations in both LLMs and MLLMs arise due to the imperfect of handling such interactions within the attention patterns of the Transformer.
\paragraph{Future directions.}
Overall, the present works and ours work reveals notable similarities in the internal patterns of hallucination between LLMs and MLLMs.
In future research, we will adopt a unified perspective to investigate the underlying causes of hallucinations in both LLMs and MLLMs.

\section{Detailed Experimental Setup}
\subsection{Detailed Settings for Findings 1} 
\label{appendix:exp1}
In the probing experiment, we utilize the pipeline proposed in the POPE~\citep{DBLP:conf/emnlp/LiDZWZW23} to construct 1,200 balanced positive and negative sample pairs from the MSCOCO dataset as training data for the probe classifier, where each sample consists of an object accompanied by a label indicating its existence or non-existence. (\textbf{Note}: There is no overlap between the training data and the evaluation data for object hallucination proposed by the POPE).
We select the AMBER dataset~\citep{amber}, which has a different distribution from the MSCOCO dataset, to test whether our conclusions can generalize. The AMBER dataset contains 1,004 carefully annotated images, each labeled with existent objects as well as non-existent objects. We use the prompt ``\texttt{Describe the image.}'' to generate raw responses from LLaVA-1.5 on the images and then extract all object category tokens and label them with whether they exist. Given that the training set contains only 80 object categories, we denote the object tokens in test data belonging to these 80 categories as in-distribution (in-dist), while the remaining tokens are categorized as out-of-distribution (OOD).

Previous work~\citep{prismatic} has demonstrated that increasing the resolution of the vision encoder enhances the visual comprehension capabilities of MLLMs. In our study, we compare LLaVA trained with a resolution of 224px against the original LLaVA with a resolution of 336px in probing experiments. Notably, the language model's weights differ between the two MLLMs, although both initial models are based on Vicuna-1.5-7b. Our results, as illustrated in the Figure~\ref{fig:arc2}, further affirm the scaling law associated with visual resolution, while also providing indirect validation of the reliability of the probing experiments.

\begin{wraptable}{r}{0.35\textwidth}
\small
\centering 
\vspace{-0.5em}
\caption{Randomly prompts.}
    \footnotesize
    \centering
    \setlength{\tabcolsep}{3mm}{
    \begin{tabular}{l}
        \toprule
        Prompts
        \\
        \midrule
        Describe the image.
        \\
        Please describe this image in detail.
        \\
        Generate a caption for this image.
        \\
        \bottomrule
    \end{tabular}
    }
    \vspace{-0.5em}
    \label{tab:prompts}
\end{wraptable}

\subsection{Detailed Settings for  Findings 2}
\label{appendix:exp2}
In the early exit experiment, we randomly select 500 images from MSOCO and use random prompts (shown in Table~\ref{tab:prompts}) to elicit raw responses from LLaVA-1.5-7b. We then extract all non-existent objects along with their corresponding preceding text. Specifically, for the sentence ``\texttt{Additionally, there is a car.}'', we extract the hallucinated object token ``\texttt{car}'' and the preceding text ``\texttt{Additionally, there is a}''. 
We re-input the preceding text into the MLLM and observe the changes in its internal state when predicting the next token. We denote that a total of $K$ preceding texts are selected, with the j-th preceding text denoted as $s^j$. 

\newpage
\section{Evaluation results in AMBER}
The AMBER image caption dataset consists of 1,004 images, each accompanied by meticulously annotated labels. These annotations include all objects present in the images, as well as some potential hallucinated objects. AMBER employs four evaluation metrics: CHAIR (the proportion of generated hallucinated objects among all objects), Cover (the coverage of generated objects against all ground truth objects), Hal (the proportion of hallucinations among all generated captions), and Cog (the overlap ratio with potential hallucinated objects). Lower values of CHAIR, Hal, and Cog indicate higher truthfulness for the MLLMs, while a higher Cover value signifies better diversity. We compare Deco with the baselines on the LLaVA-1.5-7b. The results are as shown in Table~\ref{tab:amber}. The results reveal that Deco demonstrates a significant advantage in truthfulness, although its diversity is somewhat lacking, yet remains within an acceptable range.

\begin{table*}[h]            
\centering          
\caption{Results of using \Ours on the AMBER image caption dataset with LLaVA-1.5-7b.} 
\scalebox{1.0}{
\begin{tabular}{@{}llllll@{}}           
\toprule          
\multirow{2}{*}{\textbf{Decoding}} & \multirow{2}{*}{\textbf{Method}} & \multicolumn{4}{c}{\textbf{LLaVA-1.5}}\\    
\cmidrule(lr){3-6}
  &  & $\mathrm{CHAIR}\downarrow$ & $\mathrm{Cover}\uparrow$ & $\mathrm{Hal}\downarrow$ & $\mathrm{Cog}\downarrow$ \\            
\midrule            
\multirow{2}{*}{\makecell[l]{Greedy}} & Vanilla & 8.2 & 48.9 & 34.3 & 4.0 \\      
& DoLa & 8.0 & \textbf{50.8} & 37.5 & 4.3 \\   
 & \textbf{\Ours (Ours)} & \textbf{6.6\down{1.6}} & 47.5\textbf{\downbad{1.4}} & \textbf{28.1\down{6.2}} & \textbf{2.8\down{1.2}} \\        
\cmidrule(l){1-6}      
\multirow{3}{*}{\makecell[l]{Beam Search}} & Vanilla & 7.1 & \textbf{50.7} & 32.4 & 3.8\\            
 & OPERA & 6.4 & 49.0 & 27.5 & 2.9\\           
 & \textbf{\Ours (Ours)} & \textbf{6.3\down{0.8}} & 46.8\textbf{\downbad{3.9}} & \textbf{25.1\down{7.3}} & \textbf{2.4\down{1.4}} \\          
\cmidrule(l){1-6}      
\multirow{3}{*}{\makecell[l]{Nucleus}} & Vanilla & 10.2 & 50.2 & 43.3 & 4.5\\            
 & VCD & 9.0 & \textbf{51.7} & 40.2  & 4.4\\           
 & \textbf{\Ours (Ours)} & \textbf{8.3\down{1.9}} & 48.0\textbf{\downbad{2.2}} & \textbf{37.5\down{5.8}} & \textbf{3.4\down{1.1}} \\         
\bottomrule          
\end{tabular}}
\label{tab:amber}            
\end{table*}

\section{Ablation results of Dynamic soft modulation}
\begin{figure*}[t]  
    \centering  
    \includegraphics[width=1.0\linewidth]{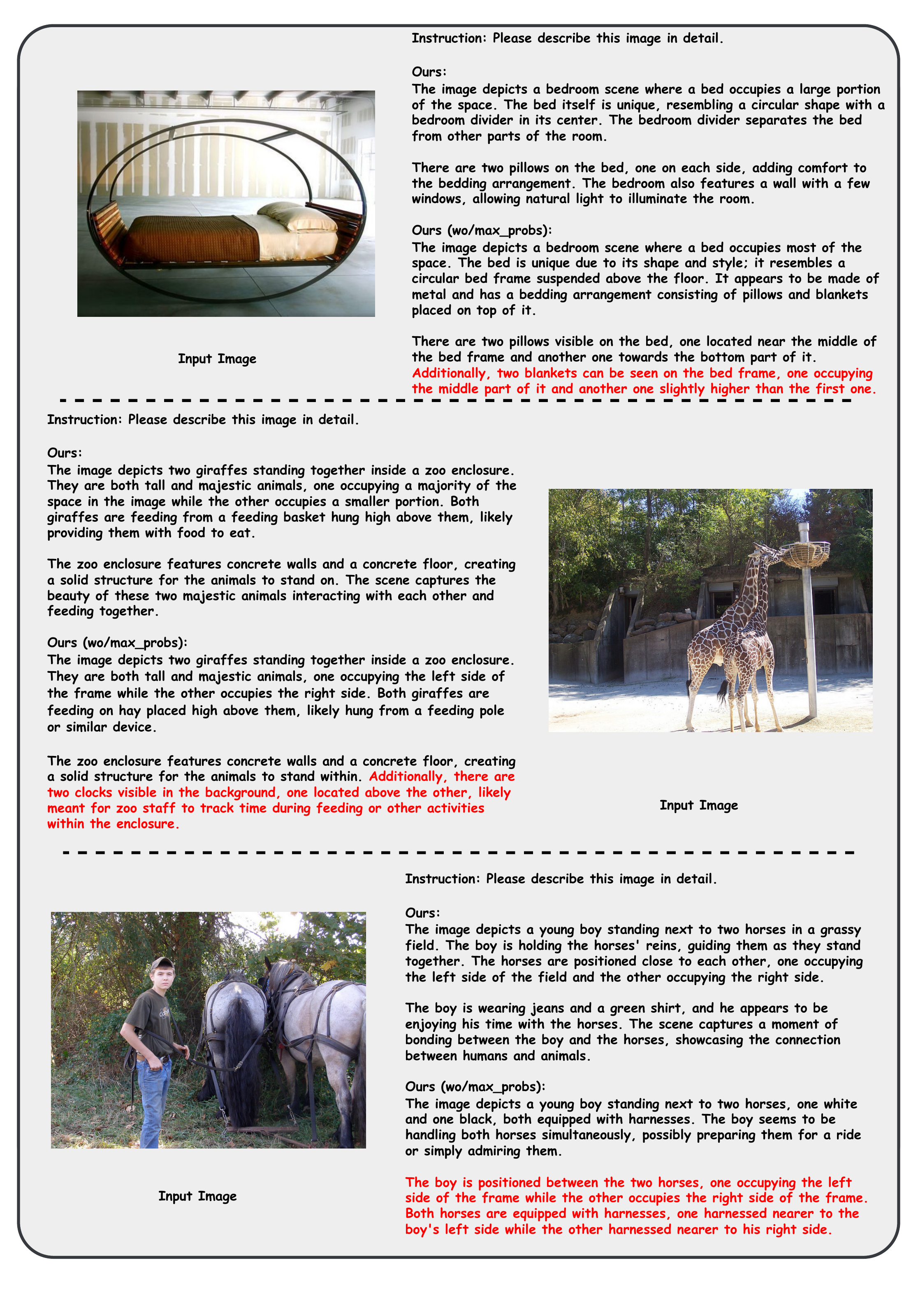}  
    \caption{Analysis of cases of soft modulation. Soft modulation avoids the hallucination phenomena and erroneous semantics caused by abrupt changes in logits.} 
    \label{fig:soft_cases}  
\end{figure*}

To quantify the effect of soft modulation, we remove the ``\texttt{max\_prob}'' term and use greedy decoding to describe the images.
The images and prompts used in the ablation experiment are consistent with the setup in Table~\ref{tab:chair}, and the ablation results are presented in the Table~\ref{tab:soft}. 
Additionally, we provide illustrative cases that demonstrate how soft modulation helps prevent abrupt changes in logits, as shown in the Figure~\ref{fig:soft_cases}.
\begin{table}[ht]
\caption{Ablation study of dynamic soft modulation.}
    \footnotesize
    \centering
    \scalebox{1.0}{
    \begin{tabular}{lllllll}
    
        \toprule
        \multicolumn{1}{c}{\multirow{2}{*}{Method}}
        & \multicolumn{2}{c}{LLaVA-1.5} 
        & \multicolumn{2}{c}{Qwen-VL} 
        \\
        \cmidrule{2-3}
        \cmidrule{4-5}
        & $\mathrm{CHAIR_S}\downarrow$ & $\mathrm{CHAIR_I}\downarrow$ & $\mathrm{CHAIR_S}\downarrow$ & $\mathrm{CHAIR_I}\downarrow$
        \\
        \midrule
        \Ours & 37.8 & 11.1 & 42.2 & 10.7
        \\
        \Ours (wo/max\_probs) & 41.2\textbf{\upbad{3.4}} & 11.6\textbf{\upbad{0.5}} & 45.8\textbf{\upbad{3.6}} & 12.3\textbf{\upbad{1.6}}
        \\
        \bottomrule
    \end{tabular}
    }
    \label{tab:soft}
\end{table}


\newpage
\section{Results of MMVet}
We evaluate the performance of LLava-1.5-7B and Qwen-VL-7B on a comprehensive benchmark MMVet and use their default settings based on nucleus sampling. The results are as shown in Table~\ref{tab:mmvet}.

\begin{table*}[h]            
\centering          
\caption{Results of using \Ours on MMVet with LLaVA-1.5-7b and Qwen-VL-7b.} 
\scalebox{0.8}{
\begin{tabular}{@{}lllllllll@{}}           
\toprule          
\textbf{Model} & \textbf{Method}  & $\mathrm{Rec}\uparrow$ & $\mathrm{OCR}\uparrow$ & $\mathrm{Know}\uparrow$ & $\mathrm{Gen}\uparrow$ & $\mathrm{Spat}\uparrow$ & $\mathrm{Math}\uparrow$ & $\mathrm{Total}\uparrow$
  \\            
\midrule            
\multirow{2}{*}{\makecell[l]{LLaVA-1.5}} & Vanilla & 28.8 & 14.1 & 15.5 & 16.4 & 15.6 & 3.5 & 23.6\\  
 & \textbf{\Ours (Ours)} & \textbf{32.1\up{3.3}} & 21.5\textbf{\up{7.4}} & \textbf{18.6\up{3.1}} & \textbf{20.7\up{4.3}} & \textbf{23.7\up{8.1}}& \textbf{11.2\up{7.7}}& \textbf{27.9\up{4.3}}\\        
\cmidrule(l){1-9}      
\multirow{3}{*}{\makecell[l]{Qwen-VL}} & Vanilla & \textbf{51.8} & 35.3 & \textbf{41.0} & 35.6 & 38.1 & 19.2 & 45.7\\       
& \textbf{\Ours (Ours)} & 50.5\downbad{1.3} & 38.2\textbf{\up{2.9}} & 38.8\downbad{2.2} & 33.8\downbad{1.8} & \textbf{41.7\up{3.6}}& \textbf{26.5\up{7.3}}& \textbf{46.3\up{0.6}} \\
\bottomrule          
\end{tabular}}
\label{tab:mmvet}            
\end{table*}


\section{Case Analysis Across Diverse MLLMs}
\label{appendix:case}

\begin{figure*}[h]  
    \centering  
    \includegraphics[width=1.0\linewidth]{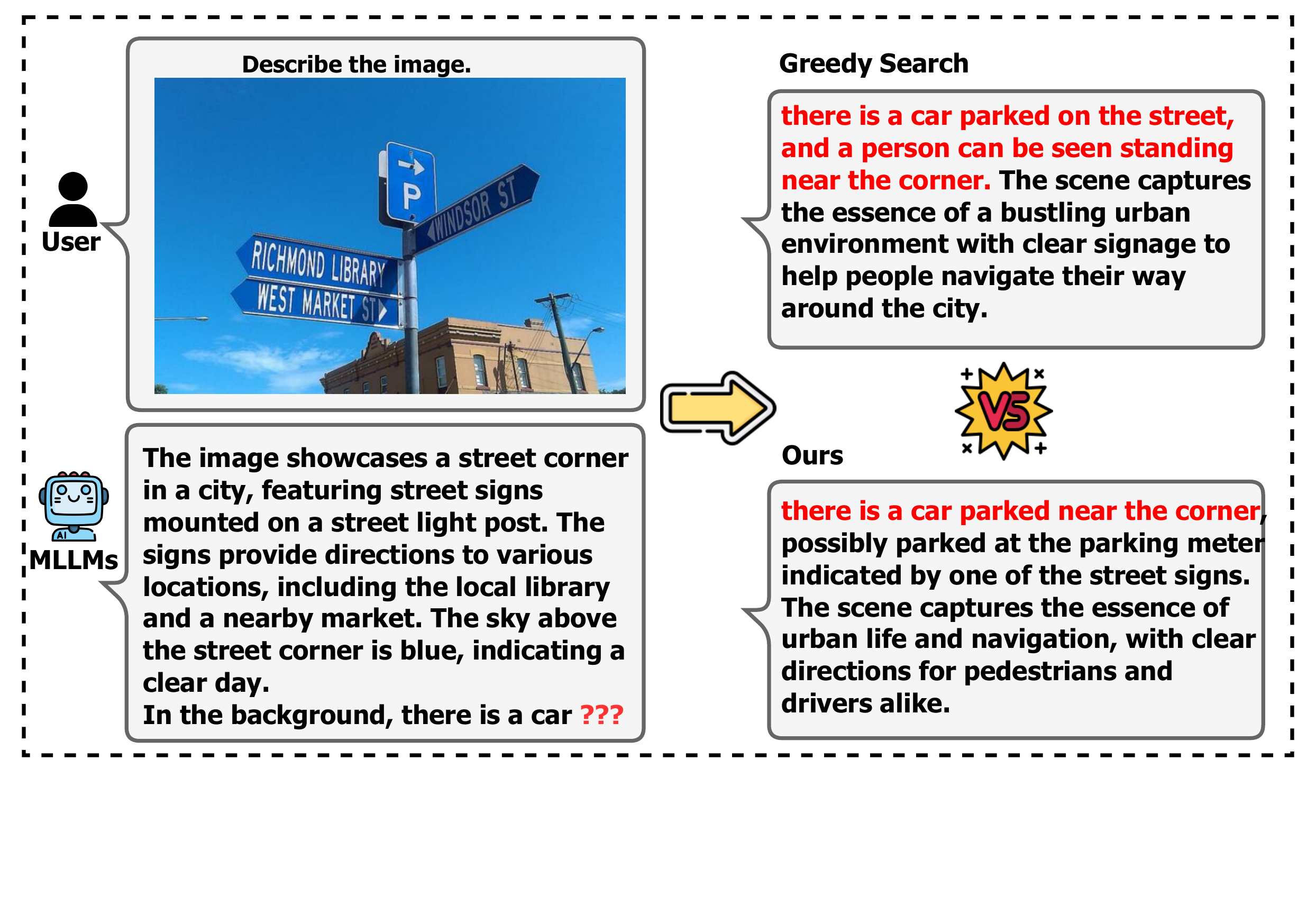}  
    \caption{The case of mitigating snowballing hallucination with \Ours.} 
    \label{snowballing}  
\end{figure*}

\begin{figure*}[t]  
    \centering  
    \includegraphics[width=1.0\linewidth]{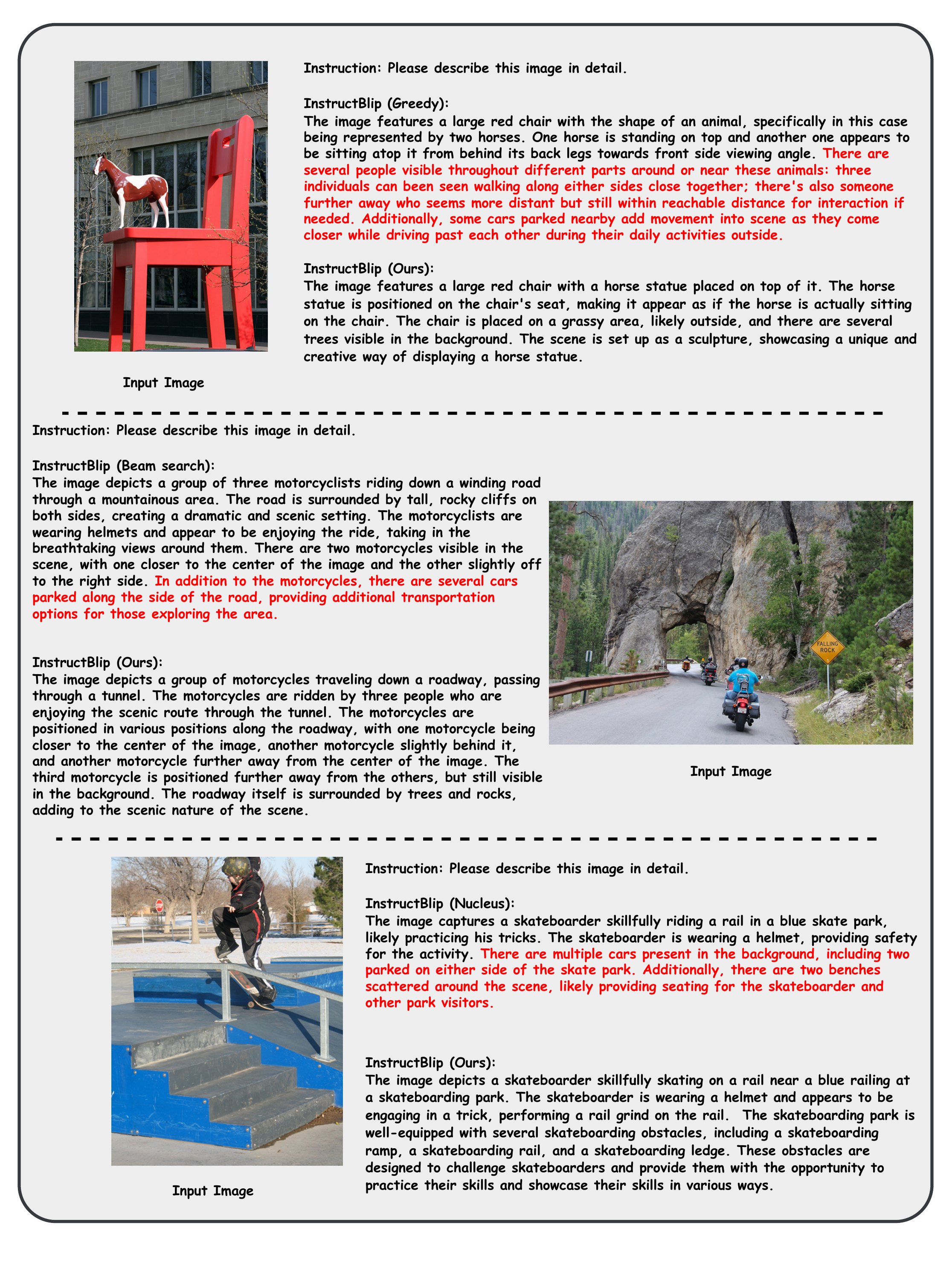}  
    \caption{\Ours's performance in reducing hallucinations of InstructBlip-7B on three basic decoing methods.} 
    \label{fig:blip_cases}  
\end{figure*}

\begin{figure*}[t]  
    \centering  
    \includegraphics[width=1.0\linewidth]{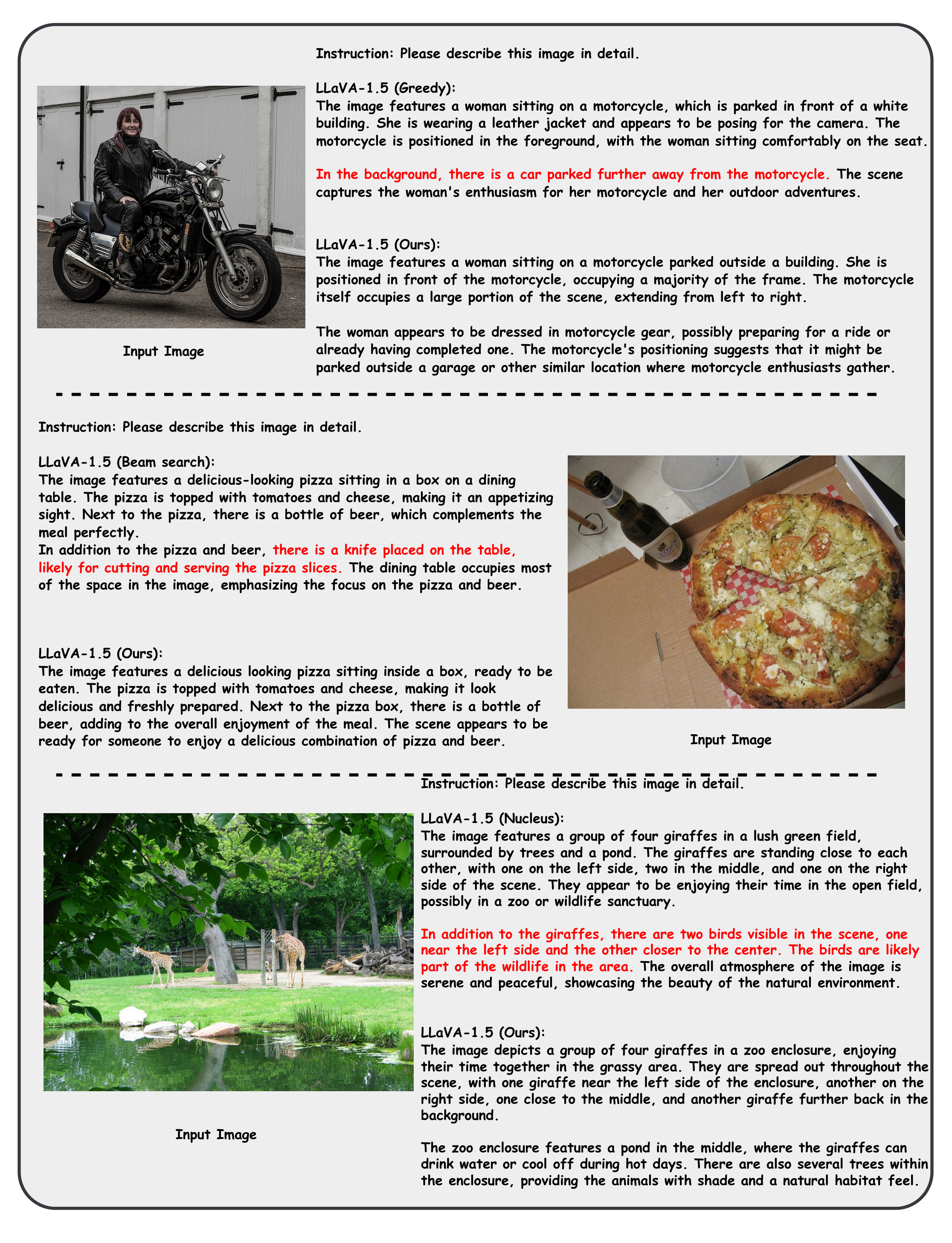}  
    \caption{\Ours's performance in reducing hallucinations of LLaVA-1.5-7B on three basic decoing methods.} 
    \label{fig:llava_cases}  
\end{figure*} 

\begin{figure*}[t]  
    \centering  
    \includegraphics[width=1.0\linewidth]{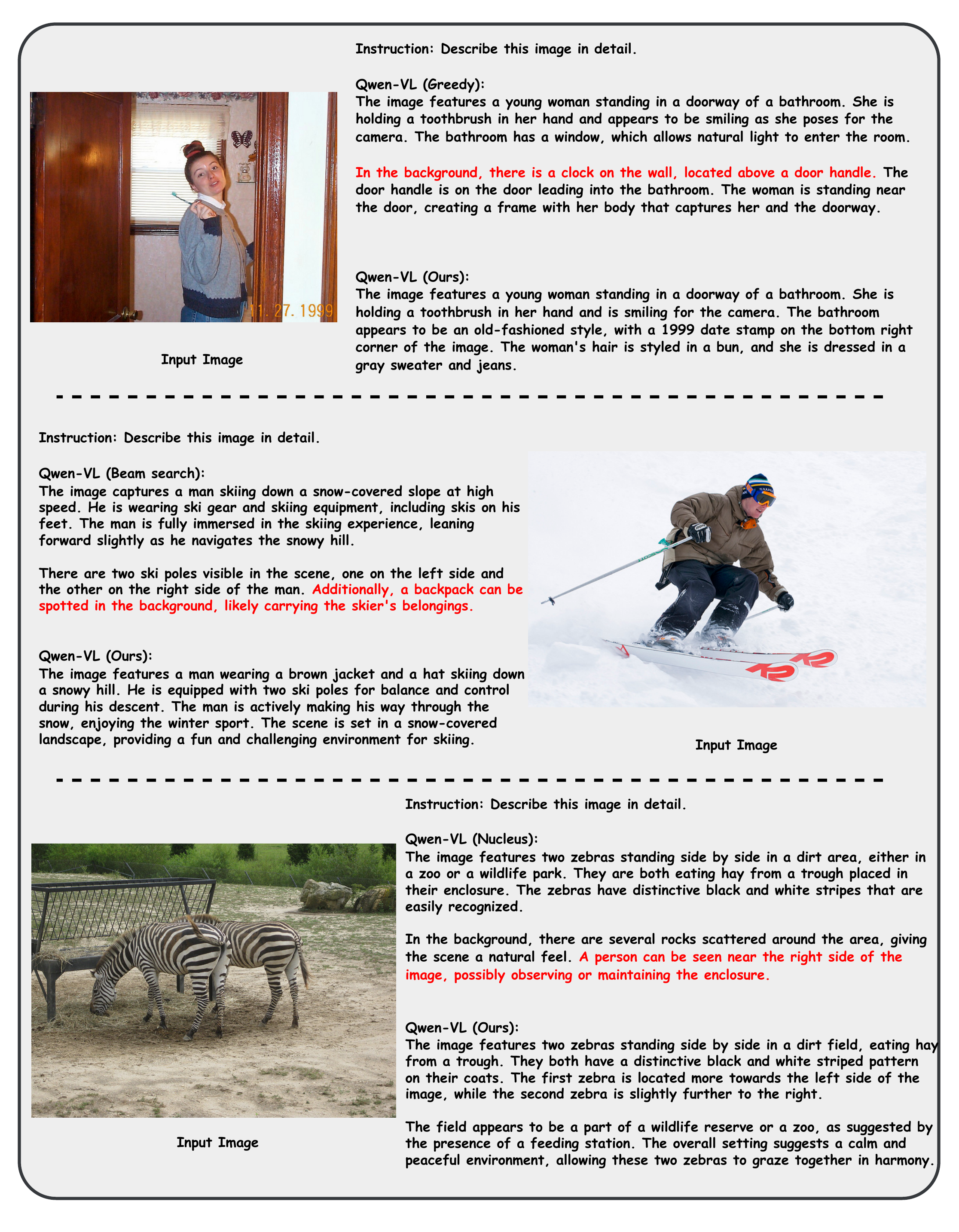}  
    \caption{\Ours's performance in reducing hallucinations of Qwen-VL-7B on three basic decoing methods.} 
    \label{fig:qwen_cases}  
\end{figure*} 

\begin{figure*}[t]  
    \centering  
    \includegraphics[width=1.0\linewidth]{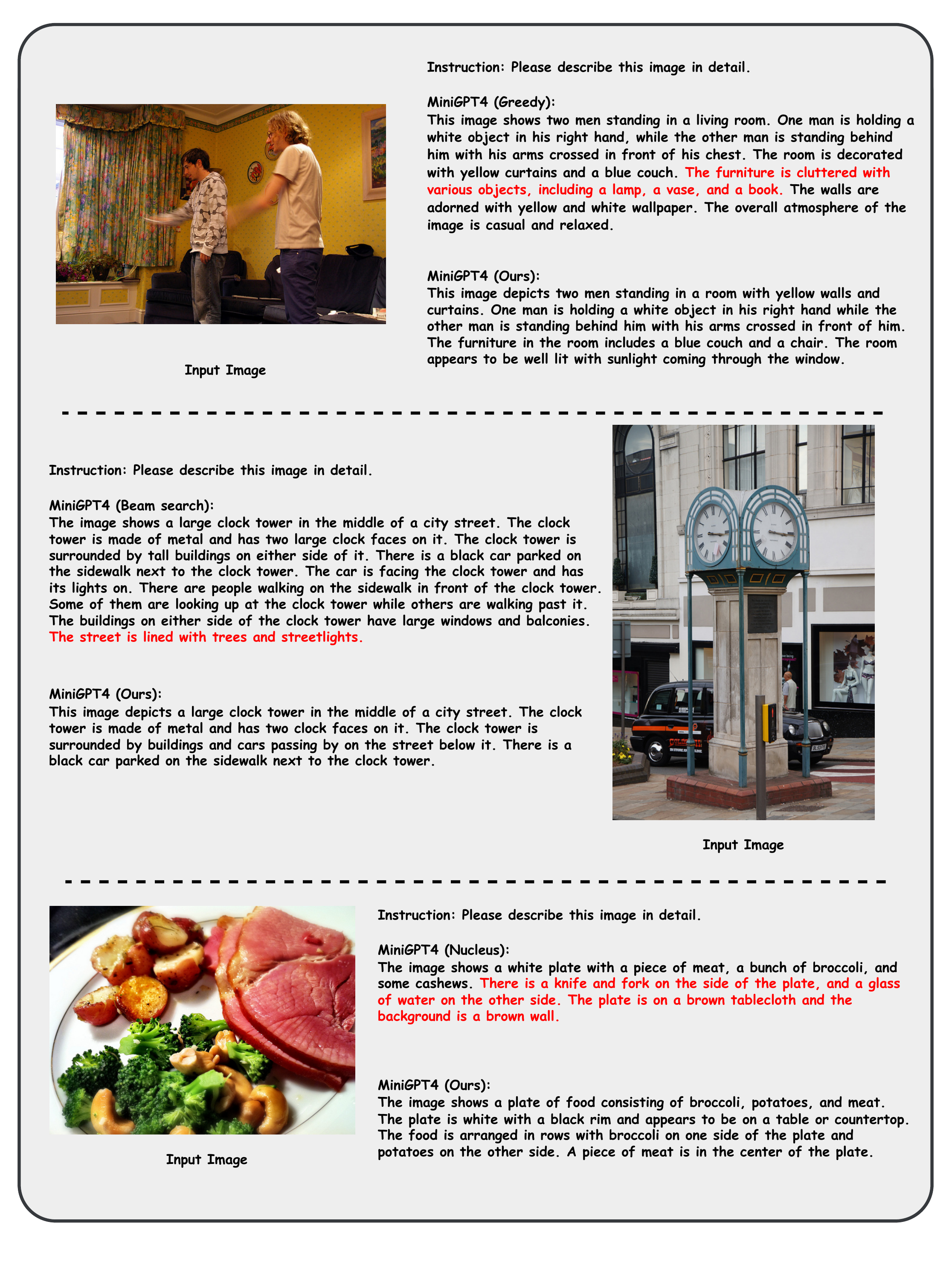}  
    \caption{\Ours's performance in reducing hallucinations of MiniGPT4-7B on three basic decoing methods.} 
    \label{fig:minigpt_cases}  
\end{figure*} 

\clearpage
\begin{table*}[h]  
\caption{The prompt used for GPT-4o evaluation adopted from \citet{vcd,opera,liu2024payingattention}}  
\label{tab:4o-prompt}  
\centering  
\begin{tabular}{p{\textwidth}}  
\hline  
\textbf{GPT-4o Prompt} \\    
\hline    
You are required to score the performance of two AI assistants in describing a given image. You should pay extra attention to the hallucination, which refers to the part of descriptions that are inconsistent with the image content, such as claiming the existence of something not present in the image or describing incorrectly in terms of the counts, positions, or colors of objects in the image. Please rate the responses of the assistants on a scale of 1 to 10, where a higher score indicates better performance, according to the following criteria: \par  
1: Accuracy: whether the response is accurate with respect to the image content. Responses with fewer hallucinations should be given higher scores. \par  
2: Detailedness: whether the response is rich in necessary details. Note that hallucinated descriptions should not count as necessary details. \par  
Please output the scores for each criterion, containing only two values indicating the scores for Assistant 1 and 2, respectively. The two scores are separated by a space. Following the scores, please provide an explanation of your evaluation, avoiding any potential bias and ensuring that the order in which the responses were presented does not affect your judgment. \par 
 \\ \par
[Assistant 1] \par  
\{Response of Assistant 1\} \par  
[End of Assistant 1] \par 
 \\ \par
[Assistant 2] \par  
\{Response of Assistant 2\} \par  
[End of Assistant 2] \par
 \\ \par
Output format: \par  
Accuracy: \textless Scores of the two answers\textgreater \par  
Reason: \par  
 \\ \par
Detailedness: \textless Scores of the two answers\textgreater \par  
Reason: \par   
 \\ \par
\hline    
\end{tabular}  
\end{table*}

\begin{table*}[h]  
\caption{The prompt used for GPT-4o to evaluate coherence.}  
\label{tab:coh}  
\centering  
\begin{tabular}{p{\textwidth}}  
\hline  
\textbf{GPT-4o Prompt} \\    
\hline    
You are required to score the coherence of two AI assistants in describing a given image. Please rate the responses of the assistants on a scale of 1 to 10, where a higher score indicates better coherence. \par
 \\ \par
[Assistant 1] \par  
\{Response of Assistant 1\} \par  
[End of Assistant 1] \par 
 \\ \par
[Assistant 2] \par  
\{Response of Assistant 2\} \par  
[End of Assistant 2] \par
 \\ \par
Output format:
Coherence: \textless Scores of the two answers \textgreater \par  
Reason: \par
 \\ \par
\hline    
\end{tabular}  
\end{table*} 

\end{document}